\documentclass[journal]{IEEEtran}
\usepackage{times}
\usepackage{latexsym}
\usepackage{multirow}
\usepackage{graphicx}
\usepackage{booktabs}
\usepackage{amsmath}
\usepackage{cite}
\usepackage{algorithmic}
\usepackage{array}
\usepackage{url}
\usepackage{xspace}
\usepackage{enumitem}
\usepackage{color}
\setitemize[1]{itemsep=0pt,partopsep=0pt,parsep=\parskip,topsep=0pt,leftmargin=10pt}

\newcommand\ourmodel{SeKnow\xspace}

\begin{document}
%
\title{End-to-End Task-Oriented Dialog Modeling with Semi-Structured Knowledge Management}

\author{Silin Gao$^{1,3}$, Ryuichi Takanobu$^2$, Antoine Bosselut$^3$, Minlie Huang$^{1\dagger}$ \\
$^1$CoAI Group, DCST, IAI, BNRIST, Tsinghua University, Beijing, China\\
$^2$Alibaba Group, Hangzhou, China\\
$^3$NLP Group, IC, EPFL, Switzerland\\
{\tt gsl16@tsinghua.org.cn, ryuichi.gxly@alibaba-inc.com,\\
antoine.bosselut@epfl.ch, aihuang@tsinghua.edu.cn}
}





\maketitle

\begin{abstract}
Current task-oriented dialog (TOD) systems mostly manage structured knowledge (e.g. databases and tables) to guide the goal-oriented conversations.
However, they fall short of handling dialogs which also involve unstructured knowledge (e.g. reviews and documents).
In this paper, we formulate a task of modeling TOD grounded on a fusion of structured and unstructured knowledge.
To address this task, we propose a TOD system with semi-structured knowledge management, \ourmodel, which extends the belief state to manage knowledge with both structured and unstructured contents.
Furthermore, we introduce two implementations of \ourmodel based on a non-pretrained sequence-to-sequence model and a pretrained language model, respectively.
Both implementations use the end-to-end manner to jointly optimize dialog modeling grounded on structured and unstructured knowledge.
We conduct experiments on a modified version of MultiWOZ 2.1 dataset, Mod-MultiWOZ 2.1, where dialogs are processed to involve semi-structured knowledge.
Experimental results show that \ourmodel has strong performances in both end-to-end dialog and intermediate knowledge management, compared to existing TOD systems and their extensions with pipeline knowledge management schemes.
\end{abstract}
\renewcommand{\thefootnote}{\fnsymbol{footnote}}
\footnotetext[0]{This paper is a further development of our prior work \cite{gao2021hyknow} accepted to ACL-IJCNLP 2021 Findings.}
\footnotetext[2]{Corresponding author.}
\renewcommand{\thefootnote}{\arabic{footnote}}


%
\IEEEpeerreviewmaketitle


\begin{figure}[t]
\centering
\includegraphics[width=1.0\columnwidth]{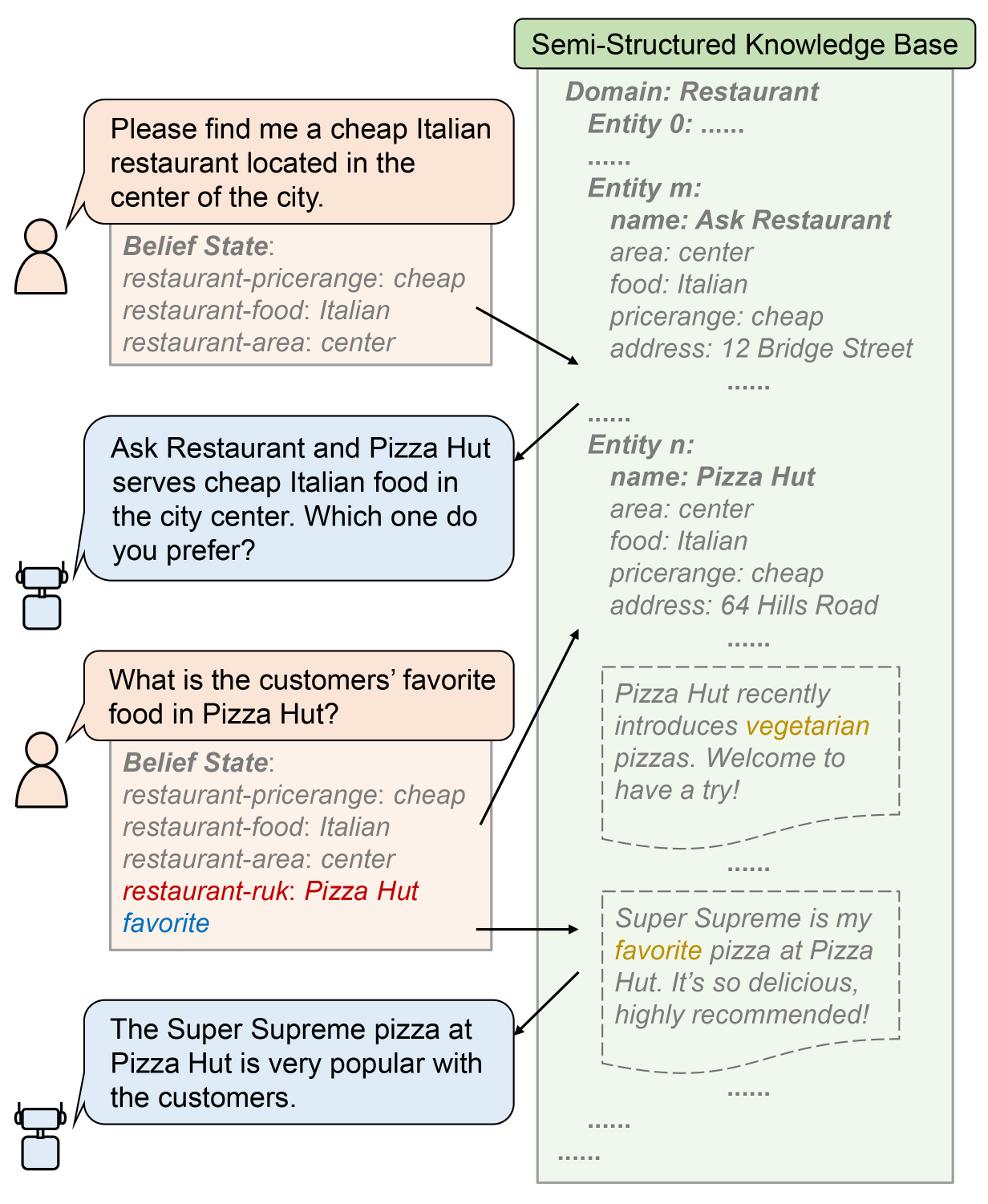}
\caption{Illustration of task-oriented dialog modeling with semi-structured knowledge management, modified from an actual sample in Mod-MultiWOZ 2.1 dataset. Words in red and blue illustrate the new domain-slot-value triple and the topic of user utterance that we introduce into the belief state, respectively. Words in yellow illustrate the topics of documents that we extract through preprocessing.}
\label{sskm}
\end{figure}

\section{Introduction}
\IEEEPARstart{R}{ecent} Task-Oriented Dialog (TOD) systems \cite{zhang2020task,le2020uniconv,zhang2020probabilistic,hosseini2020simple,peng2020soloist,kulhanek2021augpt,li2021multi} have achieved promising performance on accomplishing user goals.
Most systems typically query \textit{structured knowledge} such as tables and databases based on the user goals, and use the query results showing matched entities to guide the generation of system responses, as shown in the first dialog turn in Fig.~\ref{sskm}.

However, real-world task-oriented conversations also often involve \textit{unstructured knowledge} \cite{feng2020doc2dial} related to the user's entities of interest, such as reviews and regulation documents.
For example, as the second dialog turn in Fig.~\ref{sskm} shows, the user asks about customers' favorite food at a matched restaurant \textit{Pizza Hut}, which involves the customer reviews of this entity.
Current TOD systems fall short of handling such dialog turns since they cannot utilize relevant unstructured knowledge. This deficiency may interrupt the dialog process, causing difficulties in tracking user goals and generating system responses.

In this paper, we consider incorporating more various forms of domain knowledge into the TOD systems.
We define a task of modeling TOD which involves knowledge with both structured and unstructured contents, as shown in the semi-structured knowledge base in Fig.~\ref{sskm}.
In each dialog turn, the system needs to track the user goals associated with structured knowledge as triples and use them to query the knowledge base. The query results (i.e. the matched entities) are then used to generate the system response.
Besides, the system also needs to retrieve the unstructured contents (i.e. the documents) of knowledge base according to user goals and select relevant references (if existed) for generating the response.

To address our defined task, we propose a task-oriented dialog system with \textbf{Se}mi-Structured \textbf{Know}ledge management (\ourmodel).
It extends the belief state to handle TODs grounded on semi-structured knowledge, and further uses the extended belief state to perform both structured query and document retrieval, whose outputs are thereby used to generate the final response.
As a further development of our prior work HyKnow \cite{gao2021hyknow}, \ourmodel fuses the management of structured and unstructured knowledge via their shared domains and entities. Through the knowledge management fusion, the query of structured data can facilitate the retrieval of documents.

To investigate the trade-off between model performance and computational cost, we introduce two implementations of \ourmodel based on a non-pretrained \textbf{S}equence-\textbf{to}-\textbf{S}equence \cite{sutskever2014sequence} model (\ourmodel-S2S) and a \textbf{P}retrained \textbf{L}anguage \textbf{M}odel \cite{devlin2019bert,radford2019language} (\ourmodel-PLM), respectively.
Both implementations are in end-to-end frameworks, where dialog modeling grounded on structured and unstructured knowledge can be jointly optimized to get overall better performance.
In \ourmodel-S2S, following our prior work HyKnow \cite{gao2021hyknow}, we apply two schemes of extended belief state decoding to investigate the correlation of structured and unstructured knowledge management.
However, different from HyKnow which uses GRU \cite{cho2014properties} as backbone, we implement \ourmodel-S2S based on Universal Transformer \cite{dehghani2018universal} instead, which yields better TOD modeling performance.
In \ourmodel-PLM, we facilitate our system with large-scale pretraining models to further improve its performance, which is a brand new extension of HyKnow.

We evaluate our system on a modified version of MultiWOZ 2.1 (Mod-MultiWOZ 2.1) dataset developed from DSTC9 Track1 \cite{kim2020beyond} and original MultiWOZ 2.1 \cite{eric2020multiwoz} datasets, where dialogs involve knowledge in both structured and unstructured forms.
Experimental results show that \ourmodel-PLM and \ourmodel-S2S outperform existing TOD systems with and without model pretraining, respectively, no matter whether those TOD systems add extra components of unstructured knowledge management or not.
\ourmodel also has strong belief tracking and document retrieval performances, compared to the pipeline knowledge management schemes.

Our contributions are summarized as below:
\begin{itemize}
    \item We formulate a task of modeling TOD grounded on knowledge with both structured and unstructured contents, to incorporate more domain knowledge into the TOD systems.
    \item We propose a TOD system \ourmodel to address our proposed task, with two implementations \ourmodel-S2S and \ourmodel-PLM. Both use an extended belief state to manage semi-structured knowledge, and the end-to-end manner to jointly optimize dialog modeling grounded on structured and unstructured knowledge.
    \item Experimental results on our developed Mod-MultiWOZ 2.1 dataset show that \ourmodel has strong performance in TOD modeling grounded on semi-structured knowledge.\footnote{The code is available at \url{https://github.com/Silin159/SeKnow}}
\end{itemize}

\section{Related Work}
TOD systems usually use belief tracking, i.e. dialog state tracking (DST), to trace the user goals as \textit{belief states} through multiple dialog turns \cite{williams2013dialog,henderson2014second}.
The states are converted into a representation of constraints based on different schemes to query the databases \cite{el2017frames,budzianowski2018multiwoz,rastogi2020towards,zhu2020crosswoz}.
The entity matching results are then used to generate the system response.

To reduce deployment cost and error propagation, end-to-end trainable networks \cite{wen2017network} are introduced into TOD systems, which have continually been studied recently.
Typical end-to-end TOD systems include those with structured fusion networks \cite{mehri2019structured,le2020uniconv}, and those with multi-stage sequence-to-sequence framework \cite{lei2018sequicity,liang2020moss,zhang2020task,zhang2020probabilistic}.
With the boom of Transformers \cite{vaswani2017attention} and its large-scale pretraining \cite{devlin2019bert,radford2019language}, TOD systems based on auto-regressive language modeling have also been developed \cite{hosseini2020simple,peng2020soloist,kulhanek2021augpt}, which achieve strong TOD modeling performance.

With the development of intelligent assistants, the system should have a good command of massive external knowledge to better accomplish complicated user goals and improve user satisfaction. 
To realize this, some researchers \cite{zhao2017generative,yu2017learning,akasaki2017chat} equip the system with chatting capability to address both task and non-task content in TODs.
Other studies apply knowledge graph \cite{liao2019deep,yang2020graphdialog} or tables via SQL \cite{yu2019cosql} to enrich the knowledge of TOD systems. 
However, all these studies are still limited in dialog modeling grounded on structured knowledge.

Inspired by unstructured knowledge-grounded open-domain dialog modeling \cite{dinan2018wizard,moghe2018towards,gopalakrishnan2019topical}, there are a few studies to integrate unstructured knowledge into TOD modeling recently.
Doc2Dial \cite{feng2020doc2dial} formulates document-grounded dialog for information seeking tasks.
Beyond Domain APIs \cite{kim2020beyond} introduces knowledge snippets to answer follow-up questions out of the coverage of databases, which provides benchmarks for the 9th Dialog System Technology Challenge (DSTC9) Track-1 task \cite{kim2021beyond} and prompts some further work \cite{tan2020learning,he2021learning,thulke2021efficient,kim2021end}.
However, they only focus on dialog modeling grounded on unstructured knowledge instead.
In this paper, we aim to fill the gap of managing domain-specific knowledge with various sources and structures in end-to-end TOD systems.

\begin{figure*}[t]
\centering
\includegraphics[width=\textwidth]{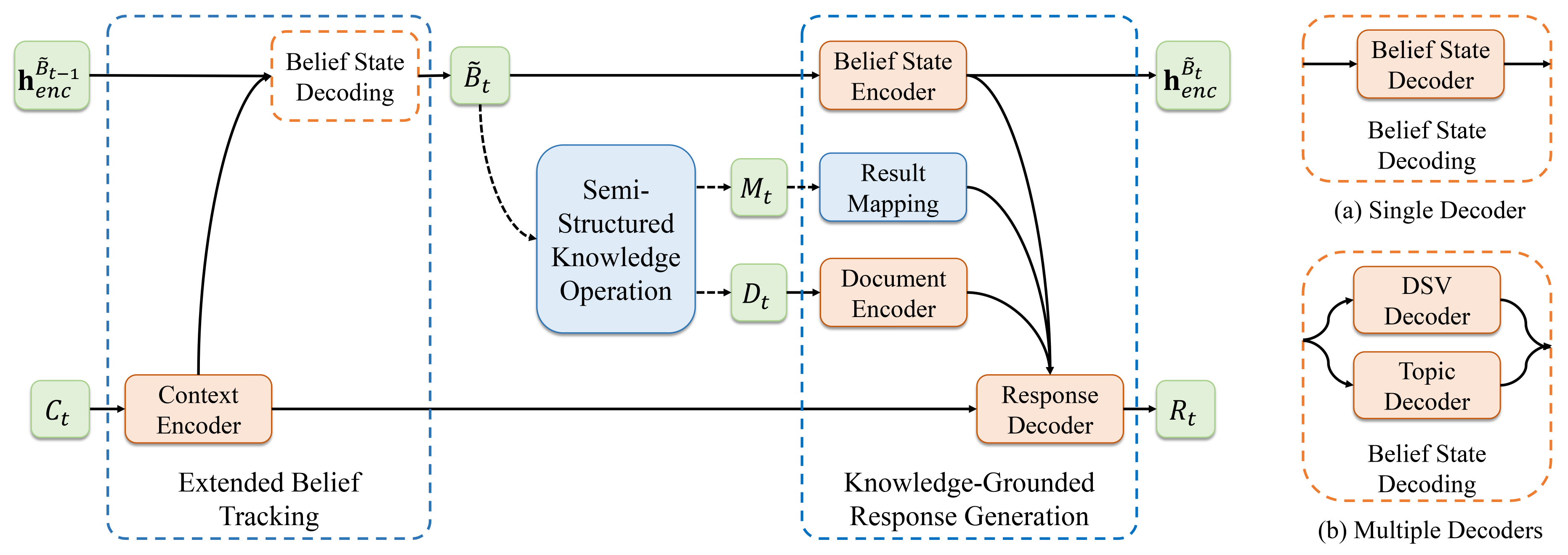}
\caption{Overview of \ourmodel-S2S.
Solid arrows denote the input/output of the encoders or decoders.
Dashed arrows denote the knowledge operation or the mapping of query result.
$C_{t}$, $M_{t}$, $D_{t}$ and $R_{t}$ represent turn $t$'s dialog context, structured query result, retrieved relevant document and system response, respectively.
$\widetilde{B}_{t}$ and $\textbf{h}_{enc}^{\widetilde{B}_{t}}$ denote the extended belief state and its hidden states at turn $t$.
The decoding of $\widetilde{B}_{t}$ (orange dashed box) is implemented in two different ways: (a) using a single decoder to generate the whole state, and (b) using two decoders to generate the domain-slot-value (DSV) triples and the topic of user utterance separately.
}
\label{seknow-s2s}
\end{figure*}

\section{Task Definition}
In this section, we introduce our formulation of modeling TOD grounded on semi-structured knowledge.
In particular, we formulate a turn-level TOD modeling task with access to a semi-structured knowledge base, which contains lists of entities characterized by different domains.
Each entity has structured attributes (e.g. name and address), and may also be associated with unstructured documents, as shown in Fig.~\ref{sskm}.
Our TOD modeling task involves both structured and unstructured contents of the knowledge base.

In each dialog turn, the system needs to track the user goals associated with structured knowledge as domain-slot-value triples in the belief state, and then query the structured contents of knowledge base to guide the generation of response.
In particular, we denote the user utterance and the system response at turn $t$ as $U_{t}$ and $R_{t}$, respectively.
Given the dialog context $C_{t}=[U_{t-k},R_{t-k},...,U_{t}]$ where $k$ is the context window size, the system needs to generate current belief state $B_{t}$ directly or by updating previously generated belief state $B_{t-1}$, which are formulated as $B_{t}=f_{b}(C_{t})$ or $B_{t}=f_{b}(C_{t},B_{t-1})$, respectively.
Then the system uses $B_{t}$ to query the structured attributes of each entity in the knowledge base, and get the query result $M_{t}$ showing matched entities, formulated as $M_{t}=f_{m}(B_{t})$.

Besides the structured query, the system also needs to retrieve the unstructured contents of knowledge base according to the user goals to find relevant references for generating the response.
Specifically, given the dialog context $C_{t}$, the system needs to retrieve the unstructured documents of each entity in the knowledge base and select a relevant document $D_{t}$ if there exists one.
The generated belief state $B_{t}$ can be optionally used to facilitate the document retrieval, formulated as $D_{t}=f_{d}(C_{t})$ or $D_{t}=f_{d}(C_{t},B_{t})$.

Finally, the system needs to generate the response $R_{t}$ based on the dialog context $C_{t}$, the belief state $B_{t}$, the structured query result $M_{t}$ and the retrieved document $D_{t}$, which is formulated as $R_{t}=f_{r}(C_{t},B_{t},M_{t},D_{t})$.

\begin{figure*}[t]
\centering
\includegraphics[width=\textwidth]{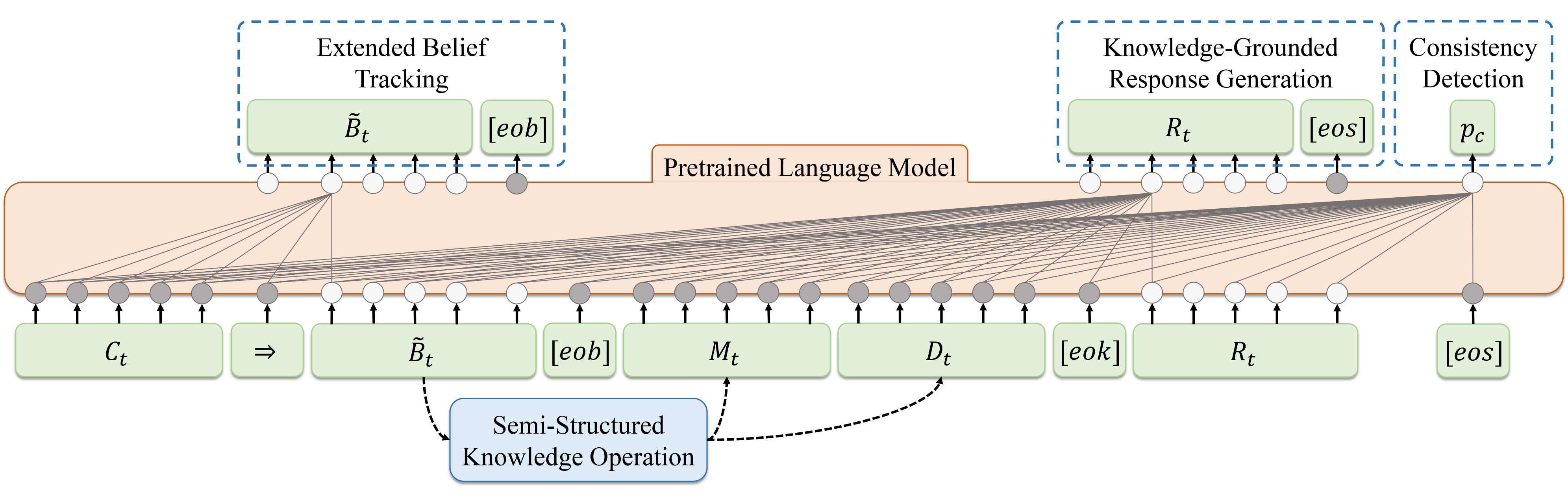}
\caption{Overview of \ourmodel-PLM.
Solid arrows denote the input/output of the pretrained language model.
Dashed arrows denote the knowledge operation.
Gray lines show the variable dependency of each language modeling sub-task.
$C_{t}$, $\widetilde{B}_{t}$, $M_{t}$, $D_{t}$ and $R_{t}$ represent turn $t$'s dialog context, extended belief state, structured query result, retrieved relevant document and system response, respectively.
$=>$, $[eob]$ are shift tokens indicating the start and end of belief tracking, while $[eok]$, $[eos]$ are shift tokens indicating the start and end of response generation.
$p_{c}$ represents the final output probability of dialog consistency.
}
\label{seknow-plm}
\end{figure*}

\section{Proposed Framework}
We propose a task-oriented dialog system with semi-structured knowledge management, \textbf{\ourmodel}, which addresses our defined task in three steps.
First, it uses \textbf{extended belief tracking} to track user goals through dialog turns that involve semi-structured knowledge.
Secondly, it performs \textbf{semi-structured knowledge operation} based on the extended belief state, to search both structured and unstructured knowledge that is relevant to the user goals.
Finally, it uses the extended belief state and relevant knowledge to perform the \textbf{knowledge-grounded response generation}.
Fig.~\ref{seknow-s2s} and \ref{seknow-plm} show our two implementations of \ourmodel based on non-pretrained sequence-to-sequence (Seq2Seq) model \cite{sutskever2014sequence} and pretrained language model (LM) \cite{devlin2019bert,radford2019language}, denoted as \textbf{\ourmodel-S2S} and \textbf{\ourmodel-PLM}, respectively.
\ourmodel-S2S is a light-weight Seq2Seq model which requires less computational resources than \ourmodel-PLM, while \ourmodel-PLM uses large-scale pretraining to yield better TOD modeling performance.

\subsection{Extended Belief Tracking}
\subsubsection{Belief State Extension}
We define an extended belief state $\widetilde{B}_{t}$ which is applicable to track user goals in TODs that involve semi-structured knowledge.
Specifically, to capture user goals associated with structured knowledge, $\widetilde{B}_{t}$ contains the domain-slot-value triples of original $B_{t}$.
While in dialog turns where user goals involve unstructured knowledge, $\widetilde{B}_{t}$ has an additional slot \textit{ruk} to indicate that current turn \textbf{r}equires \textbf{u}nstructured \textbf{k}nowledge. The prefix and value of the slot \textit{ruk} represent the involved domain and entity, e.g. \textit{restaurant}-\textit{ruk}: \textit{Pizza Hut} colored in red in Fig.~\ref{sskm}.
We denote the combination of original and newly introduced domain-slot-value triples as $DSV_{t}$.
Following the work of Sequicity \cite{lei2018sequicity}, we format $DSV_{t}$ as a text span $[DSV_{t,0},DSV_{t,1},...,DSV_{t,l^{S}_{t}-1}]$ with length $l^{S}_{t}$ to make it suitable as the input/output of our system, e.g. \textit{restaurant} \{ \textit{food} $=$ \textit{italian} $,$ \textit{area} $=$ \textit{center} \}.
Besides, the \textit{topic} of $U_{t}$ is abstracted in $\widetilde{B}_{t}$ as a word sequence $T_{t}=[T_{t,0},T_{t,1},...,T_{t,l^{T}_{t}-1}]$ with length $l^{T}_{t}$ in each turn related to unstructured knowledge, e.g. \textit{favorite} colored in blue in Fig.~\ref{sskm}.
$\widetilde{B}_{t}=[\widetilde{B}_{t,0},\widetilde{B}_{t,1},...,\widetilde{B}_{t,l^{B}_{t}-1}]$ is finally the concatenation of $DSV_{t}$ and $T_{t}$, i.e. $[DSV_{t,0},...,DSV_{t,l^{S}_{t}-1}, T_{t,0},...,T_{t,l^{T}_{t}-1}]$, where $l^{B}_{t}=l^{S}_{t}+l^{T}_{t}$.
In this paper, we define that $X_{t,0:y-1}$ denotes all tokens of $X_{t}$ before position $y$, i.e. $[X_{t,0},X_{t,1},...,X_{t,y-1}]$.

\subsubsection{Extended Belief State Decoding}\label{sec:ebsd}
Our two system implementations \ourmodel-S2S and \ourmodel-PLM decode $\widetilde{B}_{t}$ in different ways, which are described as below.

In light-weight \ourmodel-S2S, to reduce the complexity of TOD modeling, we decode $\widetilde{B}_{t}$ on the basis of previous turn's generated state $\widetilde{B}_{t-1}$.
We choose a small dialog context window where $C_{t}$ includes only previous system response $R_{t-1}$ and current user utterance $U_{t}$, because $\widetilde{B}_{t-1}$ already summarizes the information in utterances before $R_{t-1}$.
Specifically, following Seq2Seq framework, we first use the \textit{context encoder} to encode $C_{t}$, and then decode $\widetilde{B}_{t}$ based on the hidden states of context encoder $\textbf{h}_{enc}^{C_{t}}$ and previous extended belief state $\textbf{h}_{enc}^{\widetilde{B}_{t-1}}$.
Noticing that $DSV_{t}$ and $T_{t}$ are grounded on quite different vocabularies, we consider decoding $\widetilde{B}_{t}$ under two schemes: (a) using the belief state decoder to generate the whole $\widetilde{B}_{t}$, and (b) using the DSV decoder and the topic decoder to generate $DSV_{t}$ and $T_{t}$ separately.
Each implementation has its own advantages over the other.

Under the single-decoder scheme, the decoding of $DSV_{t}$ and $T_{t}$ can be jointly optimized via shared parameters:
\begin{align}
\begin{split}
    \textbf{h}_{enc}^{{C}_{t}}&=\mathrm{Encoder^{(C)}}({C}_{t}), \\
    \widetilde{B}_{t,i}&=\mathrm{Decoder^{(B)}}(\widetilde{B}_{t,0:i-1}|\textbf{h}_{enc}^{{C}_{t}},\textbf{h}_{enc}^{\widetilde{B}_{t-1}}),
\end{split}
\end{align}
where $i=0,1,...,l^{B}_{t}-1$.

While under the multi-decoder scheme, the decoding of $DSV_{t}$ and $T_{t}$ are fitted to their own smaller decoding spaces (vocabularies), and thus the generation of $\widetilde{B}_{t}$ can be decomposed into two simpler decoding processes:
\begin{align}
\begin{split}
   \textbf{h}_{enc}^{{C}_{t}}&=\mathrm{Encoder^{(C)}}({C}_{t}), \\
   DSV_{t,j}&=\mathrm{Decoder^{(S)}}(DSV_{t,0:j-1}|\textbf{h}_{enc}^{{C}_{t}},\textbf{h}_{enc}^{\widetilde{B}_{t-1}}), \\
   T_{t,k}&=\mathrm{Decoder^{(T)}}(T_{t,0:k-1}|\textbf{h}_{enc}^{{C}_{t}},\textbf{h}_{enc}^{\widetilde{B}_{t-1}}), \\
   \widetilde{B}_{t}&=[DSV_{t},T_{t}],
\end{split}
\end{align}
where $j=0,1,...,l^{S}_{t}-1$ and $k=0,1,...,l^{T}_{t}-1$.

In \ourmodel-PLM where large-scale pretraining is used to get strong TOD modeling capability, we decode $\widetilde{B}_{t}$ directly based on ${C}_{t}$ to avoid the error propagation from previous state $\widetilde{B}_{t-1}$.
Since $\widetilde{B}_{t-1}$ is not used in the decoding of $\widetilde{B}_{t}$, we choose a large context window where $C_{t}$ includes the whole dialog history, i.e. $[U_{0},R_{0},...,U_{t}]$.
We insert shift tokens $=>$ and $[eob]$ before and after $\widetilde{B}_{t}$ to indicate the start and end of belief state decoding, respectively.
Based on ${C}_{t}$, we then use the pretrained language model (PLM) to generate $\widetilde{B}_{t}$ in the left-to-right auto-regressive manner:
\begin{align}
    \widetilde{B}_{t,i}=\mathrm{PLM}({C}_{t},\widetilde{B}_{t,0:i-1}),
\end{align}
where $i=0,1,...,l^{B}_{t}-1$.

\subsection{Semi-Structured Knowledge Operation}
Based on the extended belief state $\widetilde{B}_{t}$, we conduct both structured data query and unstructured document retrieval in the knowledge base, whose outputs $M_{t}$ and $D_{t}$ are used to guide the generation of system response.
The structured query can facilitate the unstructured document retrieval by helping to identify the relevant entity and narrow down the document candidates.
This is different from our prior work HyKnow \cite{gao2021hyknow} where structured and unstructured knowledge are operated independently of each other.

We first query the knowledge base to select entities whose domains and structured attributes \textbf{exact match} the triples $DSV_{t}$ in $\widetilde{B}_{t}$.
In dialog turns that involve unstructured knowledge, the value of slot \textit{ruk} in $DSV_{t}$ serves to \textbf{fuzzy match} the name or ID of each entity, and only the best-matched entity is selected.
Similar to $DSV_{t}$, we format the query result $M_{t}$ as a text span indicating the number of matched entities in each domain, e.g. \textit{restaurant} \textit{2} \textit{match} $,$ \textit{train} \textit{no} \textit{match}.
We only need the number of matched entities instead of their specific information, because we consider generating delexicalized responses with specific slot values replaced by placeholders to improve data efficiency \cite{wen2015semantically}.

Based on the structured query result, we then perform document retrieval to find unstructured knowledge relevant to $C_{t}$.
Specifically, we first preprocess the documents of each entity in the knowledge base, to extract each document's topic as its retrieval index\footnote{
See Appendix \ref{document} for more details of the document preprocessing.}, e.g. \textit{vegetarian} and \textit{favorite} colored in yellow in Fig.~\ref{sskm}.
Then we retrieve the documents of the best-matched entity selected in the structured query to find the relevant document $D_{t}$, where we use the topic $T_{t}$ in $\widetilde{B}_{t}$ to \textbf{fuzzy match} the topic of each document and choose the best-matched one as $D_{t}$.
Noting that $D_{t}$ is set to \textit{none} if the triple with slot \textit{ruk} or the topic of user utterance is not available, i.e. unstructured knowledge is not required at turn $t$.

\subsection{Knowledge-Grounded Response Generation}
We generate the system response ${R}_{t}$, i.e. a word sequence $[R_{t,0},R_{t,1},...,R_{t,l^{R}_{t}-1}]$ with length $l^{R}_{t}$, based on the dialog context $C_{t}$, the extended belief state $\widetilde{B}_{t}$, and the outputs of semi-structured knowledge operation $M_{t}$ and $D_{t}$.
Our two system implementations \ourmodel-S2S and \ourmodel-PLM also decode ${R}_{t}$ in different ways, which are described as below.

In \ourmodel-S2S, we first use the same context encoder in Sec.~\ref{sec:ebsd} to encode $C_{t}$ into hidden states $\textbf{h}_{enc}^{C_{t}}$.
Besides, we use the \textit{belief state encoder} and the \textit{document encoder} to encode $\widetilde{B}_{t}$ and $D_{t}$ into hidden states $\textbf{h}_{enc}^{\widetilde{B}_{t}}$ and $\textbf{h}_{enc}^{D_{t}}$, respectively.
For the structured query result $M_{t}$, we follow MultiWOZ \cite{budzianowski2018multiwoz} to map it to a vector $\textbf{m}_{t}$ according to the number of matched entities in turn $t$'s active domain and whether the booking is available.
Based on the hidden states of all the encoders and the vector $\textbf{m}_{t}$, we use the \textit{response decoder} to generate $R_{t}$, formulated as:
\begin{align}
\begin{split}
    \textbf{h}_{enc}^{\widetilde{B}_{t}}&=\mathrm{Encoder^{(B)}}(\widetilde{B}_{t}), \\ \textbf{h}_{enc}^{D_{t}}&=\mathrm{Encoder^{(D)}}(D_{t}), \\
    \textbf{m}_{t}&=\mathrm{Mapping}(M_{t}), \\
    R_{t,i}&=\mathrm{Decoder^{(R)}}(R_{t,0:i-1}|\textbf{h}_{enc}^{C_{t}},\textbf{h}_{enc}^{\widetilde{B}_{t}},\textbf{h}_{enc}^{D_{t}},\textbf{m}_{t}),
\end{split}
\end{align}
where $i=0,1,...,l^{R}_{t}-1$.

In \ourmodel-PLM, we insert shift tokens $[eok]$ and $[eos]$ before and after ${R}_{t}$ to indicate the start and end of response decoding, respectively.
Based on the concatenation of $C_{t}$, $\widetilde{B}_{t}$, $M_{t}$ and $D_{t}$, we use the pretrained language model (PLM) to generate ${R}_{t}$ in the left-to-right auto-regressive manner:
\begin{align}
    {R}_{t,i}&=\mathrm{PLM}({C}_{t},\widetilde{B}_{t},M_{t},D_{t},{R}_{t,0:i-1}),
\end{align}
where $i=0,1,...,l^{R}_{t}-1$.

\subsection{Consistency Detection}
In \ourmodel-PLM, we also consider an auxiliary consistency detection task \cite{devlin2019bert,peng2020soloist} for model training.
Specifically, we follow AuGPT \cite{kulhanek2021augpt} to randomly corrupt half of the dialog samples and train our model to detect each sample's consistency.
The corruption includes three types happening with equal probability: (a) $\widetilde{B}_{t}$ is wholly replaced by another, (b) The value of each slot in $\widetilde{B}_{t}$ is replaced by a different one. (c) ${R}_{t}$ is replaced by another.
We apply a binary classifier on the output hidden state of final token $[eos]$ to predict the probability of dialog consistency $p_{c}$, formulated as:
\begin{align}
    {p}_{c}&=\mathrm{PLM}({C}_{t},\widetilde{B}_{t},M_{t},D_{t},{R}_{t}).
\end{align}

\subsection{Model Training and Implementation Details}
In \ourmodel-S2S, we use Universal Transformers (UT) \cite{dehghani2018universal} to implement our encoders and decoders.
Unlike the original Transformers \cite{vaswani2017attention} which stacks multiple layers, UT builds the encoder/decoder as a single self-attention \& feed-forward layer with recurrent connection, combining a two-dimensional (position, time) coordinate embeddings.
Besides, UT improves the feed-forward network with depth-wise separable convolutions \cite{chollet2017xception} instead of linear transformations.
As a generalization of Transformers, UT remedies the deficiencies of Transformers in some Seq2Seq tasks, especially the copy \cite{gu2016incorporating} of text which plays a significant role in TOD modeling \cite{lei2018sequicity}.

In \ourmodel-PLM, we use pretrained GPT-2 \cite{radford2019language} as our model backbone.
Besides, we follow AuGPT \cite{kulhanek2021augpt} to further pretrain GPT-2 on large-scale TOD corpus Taskmaster-1 \cite{byrne2019taskmaster} and Schema-Guided Dialogue \cite{rastogi2020towards}.

Both \ourmodel-S2S and \ourmodel-PLM are optimized via supervised training.
In particular, each dialog turn in the training data is initially labeled with the original belief state and the relevant document.
We extend the belief state label based on the domain, entity and extracted topic of the relevant document.
Then the extended belief state label and the reference response are used to calculate the cross-entropy loss with the generated $\widetilde{B}_{t}$ and $R_{t}$, respectively, formulated as:
\begin{align}
    \mathcal{L}_{B}&=-\mathrm{log} p(\widetilde{B}_{t}|{C}_{t})=-\sum_{i=0}^{l^{B}_{t}}\mathrm{log}p(\widetilde{B}_{t,i}|{C}_{t},\widetilde{B}_{t,0:i-1}). \\
\begin{split}
    \mathcal{L}_{R}&=-\mathrm{log}p({R}_{t}|{C}_{t},\widetilde{B}_{t},{M}_{t},{D}_{t}) \\
    &=-\sum_{i=0}^{l^{R}_{t}}\mathrm{log}p({R}_{t,i}|{C}_{t},\widetilde{B}_{t},{M}_{t},{D}_{t},{R}_{t,0:i-1}).
\end{split}
\end{align}

In \ourmodel-PLM whose training contains auxiliary consistency detection task, we also calculate the binary cross-entropy loss on the output probability of dialog consistency $p_{c}$:
\begin{align}
    \mathcal{L}_{C}&=-y_{c}\mathrm{log}p_{c}-(1-y_{c})\mathrm{log}(1-p_{c}).
\end{align}
where $y_{c}$ is the label indicating whether the contents of current dialog sample are consistent ($y_{c}=1$) or not ($y_{c}=0$).

We sum all losses together and perform gradient descent in each turn to optimize the model parameters.\footnote{See Appendix \ref{implement} for more model training and implementation details.}

\section{Experimental Settings}
\subsection{Dataset}
We evaluate our proposed system on a modified version of MultiWOZ 2.1 dataset, abbreviated as Mod-MultiWOZ 2.1, which is developed based on the DSTC9 Track1 \cite{kim2020beyond} dataset and the original MultiWOZ 2.1 \cite{eric2020multiwoz} dataset.
In DSTC9 Track1 dataset, crowd-sourcing workers are hired to insert additional turns into the original MultiWOZ dialogs.
Each newly inserted turn involves unstructured knowledge, represented as a relevant document annotated with it.
All documents are characterized by different MultiWOZ domains and entities in an additional document base.
We combine the DSTC9 Track1 dataset with the original MultiWOZ 2.1 to build our Mod-MultiWOZ 2.1 dataset, in detail:
\begin{itemize}
    \item We fuse the unstructured document base of DSTC9 Track1 dataset and the structured database of original MultiWOZ 2.1 dataset, to get our semi-structured knowledge base. Specifically, for domains restaurant and hotel, we gather documents that belong to the same entity and link them to the database entry whose \textit{name} attribute matches the entity name. While for domains train and taxi, all documents belong to the same wildcard entity “*”, so we just link them to every database entry in the same domain. Note that the other three MultiWOZ domains (attraction, hospital and police) are not involved in the document base, so database entries in these three domains are not linked to any document.
    \item We recover the belief state labels which are not originally provided in the DSTC9 Track1 dataset. Specifically, we match the concatenation of utterances of each DSTC9 Track1 dialog with that of each original MultiWOZ 2.1 dialog. Each DSTC9 Track1 dialog is paired with its best-matched dialog in original MultiWOZ 2.1, where the belief state labels can be found and recovered.
    \item We split the training, development and test sets according to the original MultiWOZ 2.1, which is different from the data split in DSTC9 Track1.\footnote{See Appendix C for details of data statistics.}
\end{itemize}

\subsection{Baselines}
We compare \ourmodel with \textbf{1) existing end-to-end (E2E) TOD models and dialog state tracking (DST) models}, to explore the benefits of incorporating unstructured knowledge management into TOD modeling.
We also compare \ourmodel with \textbf{2) unstructured knowledge management models}, to investigate our system's document retrieval performance.
For the comparison with pipeline systems which have both structured and unstructured knowledge management, we also consider \textbf{3) the combinations of 1) and 2)} as our baselines.

\subsubsection{E2E TOD Models and DST Models}
We consider two light-weight baseline E2E TOD models with different types of structures:
\textbf{UniConv} \cite{le2020uniconv} uses a structured fusion \cite{mehri2019structured} design, while \textbf{LABES-S2S} \cite{zhang2020probabilistic} is based on a multi-stage Seq2Seq \cite{lei2018sequicity} architecture.
Besides, we consider two large-scale baseline E2E TOD models developed from pretrained language models:
\textbf{MinTL} (BART-large) \cite{lin2020mintl} finetunes BART \cite{lewis2020bart} to model TOD in the Seq2Seq architecture, while \textbf{AuGPT} \cite{kulhanek2021augpt} finetunes GPT-2 \cite{radford2019language} to model TOD in the auto-regressive manner, with further pretraining on large TOD corpus and training data augmentation based on back-translation \cite{sennrich2016improving,edunov2018understanding,federmann2019multilingual}.
All four E2E TOD models only manage structured knowledge (database) in their TOD modeling.
In addition to E2E TOD models, we also compare \ourmodel with existing DST models in the belief tracking evaluation.
Specifically, we use \textbf{TRADE} \cite{wu2019transferable} and \textbf{TripPy} \cite{heck2020trippy} as two DST baselines, which are representative BERT-free and BERT-based \cite{devlin2019bert} DST models, respectively.

\subsubsection{Unstructured Knowledge Management Models}
We first compare \ourmodel with \textbf{Beyond Domain APIs (BDA)} \cite{kim2020beyond}, which uses two classification modules based on BERT \cite{devlin2019bert} to detect dialog turns requiring unstructured knowledge and retrieve relevant documents, respectively.
We also compare our system with three representative models developed from BDA in the first track of the 9th Dialog System Technology Challenge (DSTC9).
Thulke et al. \cite{thulke2021efficient} propose two models based on RoBERTa \cite{liu2019roberta} to better address the document retrieval problem in BDA:
\textbf{Hierarchical Knowledge Selection (HKS)} model narrows down the candidates of document retrieval by first identifying the relevant domain and entity, while \textbf{Dense Knowledge Retrieval (DKR)} model improves the efficiency of document retrieval by formulating it as a metric learning problem.
Kim et al. \cite{kim2021end} propose an \textbf{End-to-End Document-Grounded Conversation (E2E-DGC)} model based on T5 \cite{raffel2020exploring} to optimize the document retrieval jointly with the knowledge-grounded response generation, which also yields better performance than BDA.
Moreover, we use standard information retrieval (IR) systems \textbf{TF-IDF} \cite{manning2008introduction} and \textbf{BM25} \cite{robertson2009probabilistic} as the other two baseline models.

\subsubsection{Combinations}
We combine the unstructured knowledge management model BDA or HKS with every DST or E2E TOD model.
Specifically, BDA or HKS detects dialog turns involving unstructured knowledge, and generates responses in these turns based on the dialog context and retrieved documents.
While the DST or E2E TOD model handles the rest dialog turns which are only related to structured knowledge.
Noting that BDA uses finetuned GPT-2 \cite{radford2019language} to generate responses, while HKS finetunes BART \cite{lewis2020bart} instead and follow REALM \cite{guu2020realm} to apply retrieval augmented response generation.


Noting that DST and E2E TOD models based on BERT, BART or GPT-2 utilizes large-scale language model (LM) pretraining to improve their TOD modeling performance, which however requires large model sizes and computing resources.
For fair comparisons, we distinguish them from other non-pretrained light-weight models in our experiments.

\section{Results and Analysis}
We compare \ourmodel-S2S with light-weight non-pretrained baseline models, and test its performance under both the single-decoder and multi-decoder belief state decoding schemes, denoted as \ourmodel-S2S (Single) and \ourmodel-S2S (Multiple), respectively.
Besides, we compare \ourmodel-PLM with baseline models which utilize large-scale pretrained language models (LM).
All implementations of \ourmodel come to the same conclusions when compared with their corresponding baseline systems, which are described in detail below.

\begin{table*}[t]
\caption{End-to-end evaluation results on Mod-MultiWOZ 2.1. ``+" denotes the model combination. Best results among systems with and without pretrained language models (LM) (i.e. below and above the internal dividing line) are separately marked in bold.}
\label{table1}
\centering
\resizebox{0.92\textwidth}{!}{
\smallskip\begin{tabular}{lccccccc}
\hline
Model               & Pretrained LM &     Inform    &    Success    &      BLEU     &     METEOR    &     ROUGE-L   &    Combined   \\
\hline
UniConv             &      none     &      71.5     &      61.8     &      18.5     &      37.8     &      40.5     &      85.7     \\
UniConv + BDA       &        -      &      72.0     &      62.6     &      16.9     &      35.7     &      38.9     &      84.2     \\
UniConv + HKS       &        -      &      72.8     &      63.5     &      17.9     &      37.2     &      39.7     &      86.1     \\
LABES-S2S           &      none     &      76.5     &      65.3     &      17.8     &      36.8     &      39.9     &      88.7     \\
LABES-S2S + BDA     &        -      &      77.1     &      66.2     &      15.7     &      33.8     &      37.8     &      87.4     \\
LABES-S2S + HKS     &        -      &      78.2     &      66.7     &      18.2     &      37.5     &      40.1     &      90.7     \\
\ourmodel-S2S (Single)   &   none   & \textbf{82.9} & \textbf{68.7} & \textbf{19.0} & \textbf{38.6} & \textbf{40.8} & \textbf{94.8} \\
\ourmodel-S2S (Multiple) &   none   &      80.6     &      68.4     &      18.7     &      38.1     &      40.3     &      93.2     \\
\hline
MinTL               &      BART     &      82.1     &      70.2     &      16.9     &      36.1     &      39.1     &      93.1      \\
MinTL + BDA         &        -      &      83.8     &      71.5     &      16.8     &      35.6     &      39.3     &      94.5      \\
MinTL + HKS         &        -      &      83.9     &      71.7     &      17.1     &      36.5     &      39.4     &      94.9      \\
AuGPT               &      GPT-2    &      88.9     &      69.5     &      16.8     &      36.1     &      39.3     &      96.0      \\
AuGPT + BDA         &        -      &      91.2     &      70.4     &      16.8     &      36.0     &      39.1     &      97.6      \\
AuGPT + HKS         &        -      &      91.6     &      70.7     &      17.0     &      36.3     &      39.3     &      98.2      \\
\ourmodel-PLM       &      GPT-2    & \textbf{93.6} & \textbf{71.9} & \textbf{17.3} & \textbf{36.8} & \textbf{40.0} & \textbf{100.1} \\
\hline
\end{tabular}
}
\end{table*}

\subsection{End-to-End Evaluation}\label{sec:e2e}
Table \ref{table1} shows our experimental results of the end-to-end (E2E) TOD evaluation, where we evaluate the task completion rate and language quality of system responses.
In terms of the task completion rate, we measure whether the system provides correct entities (\textbf{Inform} rate) and answers all the requested information (\textbf{Success} rate) in a dialog, following MultiWOZ \cite{budzianowski2018multiwoz}.
For the evaluation of language quality, we adopt commonly used metrics \textbf{BLEU} \cite{papineni2002bleu}, \textbf{METEOR} \cite{banerjee2005meteor} and \textbf{ROUGE-L} \cite{lin2004rouge}.
Moreover, we use \textbf{Combined} score computed by $(Inform+Success)\times0.5+BLEU$ for overall evaluation, as suggested by MultiWOZ 2.1 \cite{eric2020multiwoz}.

We find that \ourmodel-S2S has significantly better task completion rate compared to the non-pretrained E2E TOD models.
It also generates responses with better language quality than the E2E TOD models.
The same conclusions are also observed by comparing \ourmodel-PLM with pretrained LM based E2E TOD models.
All above results show that our belief state extension helps to distinguish whether a dialog turn does or does not involve unstructured knowledge, which avoids the confusion between handling the two kinds of dialog turns.
In addition, we retrieve unstructured documents to provide relevant references for generating the response, which guide our system to give more appropriate responses in dialog turns that involve unstructured knowledge.

We also observe that \ourmodel-S2S and \ourmodel-PLM outperform the combinations of BDA/HKS with non-pretrained and pretrained E2E TOD models, respectively.
This indicates that our end-to-end model framework has advantages over the pipeline structures of combination models.
In particular, dialog modeling grounded on the structured and unstructured knowledge are integrated in a uniform architecture in our system, where they are jointly optimized to an overall better performance.
Since our system is trained end-to-end, it also has lower deployment cost in real-world applications compared to the pipeline systems.

\begin{table}[t]
\caption{Context-to-response generation on Mod-MultiWOZ 2.1. All symbols and markings have same meaning as in Table \ref{table1}.}
\label{table2}
\centering
\resizebox{1.0\columnwidth}{!}{
\smallskip\begin{tabular}{l@{~~}c@{~~}c@{~~}c@{~~}c}
\hline
Model               &     Inform    &    Success    &      BLEU     &    Combined    \\
\hline
UniConv             &      84.2     &      71.8     &      19.0     &      97.3      \\
UniConv + HKS       &      85.8     &      73.9     & \textbf{19.5} &      99.6      \\
LABES-S2S           &      83.6     &      74.2     &      18.3     &      97.2      \\
LABES-S2S + HKS     &      85.0     &      75.3     &      19.2     &      99.4      \\
\ourmodel-S2S       & \textbf{87.6} & \textbf{76.8} & \textbf{19.5} & \textbf{101.7} \\
\hline
MinTL               &      91.7     &      77.4     &      17.4     &      102.0      \\
MinTL + HKS     &      92.5     &      77.8     &      17.7     &      102.9     \\
AuGPT               &      94.2     &      76.4     &      17.4     &      102.7     \\
AuGPT + HKS         &      95.3     &      77.6     &      17.5     &      104.0     \\
\ourmodel-PLM       & \textbf{96.0} & \textbf{78.0} & \textbf{17.9} & \textbf{104.9} \\
\hline
\end{tabular}
}
\end{table}

\subsection{Context-to-Response Generation}
We also conduct evaluations on the context-to-response (C2R) generation, where systems directly use the oracle belief state and knowledge to generate the response.
The experimental results are shown in Table \ref{table2}, where we observe the same conclusions as in the E2E evaluation (Table \ref{table1}). This again shows our system's superiority in TOD modeling grounded on semi-structured knowledge.
Noting that we do not differentiate the two belief state decoding schemes of \ourmodel-S2S because they share the same substructure in the C2R task.
We also do not list the combination models with BDA because their performances are almost the same as those with HKS.

Additionally, we observe that \ourmodel's performance gap between E2E and C2R evaluations is smaller than the baseline models, reflected in the smaller variations of the combined score.
This shows that the belief state and knowledge provided by our system are probably closer to the oracle and may give stronger guidance to generate the response.

\begin{table}[t]
\caption{Original turns' belief tracking results on Mod-MultiWOZ 2.1. All symbols and markings have same meaning as in Table \ref{table1}.}
\label{table3}
\centering
\resizebox{0.98\columnwidth}{!}{
\smallskip\begin{tabular}{lcc}
\hline
Model               & Pretrained LM &   Joint Goal  \\
\hline
TRADE               &      none     &      42.9     \\
TRADE + BDA         &        -      &      43.8     \\
TRADE + HKS         &        -      &      43.9     \\
UniConv             &      none     &      45.5     \\
UniConv + BDA       &        -      &      46.5     \\
UniConv + HKS       &        -      &      47.0     \\
LABES-S2S           &      none     &      46.0     \\
LABES-S2S + BDA     &        -      &      46.8     \\
LABES-S2S + HKS     &        -      &      47.6     \\
\ourmodel-S2S (Single)   &      none     & \textbf{49.1} \\
\ourmodel-S2S (Multiple) &      none     &      48.4     \\
\hline
TripPy              &      BERT     &      50.4     \\
TripPy + BDA        &        -      &      51.2     \\
TripPy + HKS        &        -      &      51.4     \\
MinTL               &      BART     &      52.0     \\
MinTL + BDA         &        -      &      53.2     \\
MinTL + HKS         &        -      &      53.5     \\
AuGPT               &      GPT-2    &      55.0     \\
AuGPT + BDA         &        -      &      56.0     \\
AuGPT + HKS         &        -      &      56.6     \\
\ourmodel-PLM       &      GPT-2    & \textbf{58.5} \\
\hline
\end{tabular}
}
\end{table}

\begin{table}[t]
\caption{Document retrieval results on Mod-MultiWOZ 2.1. Best results are marked in bold.}
\label{table4}
\centering
\resizebox{1.0\columnwidth}{!}{
\smallskip\begin{tabular}{l@{~~}c@{~~~}c@{~~}c}
\hline
Model               &       Type       &     MRR@5     &     R@1       \\
\hline
TF-IDF              &    standard IR   &     68.7      &     54.1      \\
BM25                &    standard IR   &     69.2      &     52.5      \\
BDA                 &  classification  &     80.6      &     69.8      \\
E2E-DGC             &  classification  &     89.9      &     87.1      \\
DKR                 &  dense retrieval &     92.9      &     89.8      \\
HKS                 &  classification  & \textbf{95.7} &     91.9      \\
\hline
\ourmodel-S2S (Single)   & topic match &     91.4      &     89.9      \\
\ourmodel-S2S (Multiple) & topic match &     90.6      &     89.0      \\
\ourmodel-PLM            & topic match &     94.7      & \textbf{93.4} \\
\hline
\end{tabular}
}
\end{table}

\subsection{Knowledge Management Evaluation}\label{sec:km}
To further investigate our system's E2E performance, we conduct evaluations on the intermediate structured and unstructured knowledge management.
In terms of structured knowledge management, we evaluate the belief tracking performance which directly determines the structured data query accuracy.
Specifically, we use the \textbf{Joint Goal} accuracy \cite{henderson2014second} to measure whether belief states are predicted correctly in the original dialog turns of Mod-MultiWOZ 2.1.
Noting that we do not consider the newly inserted dialog turns where belief states are not uniformly defined: \ourmodel uses the extended belief state, while baseline DST/E2E models only parse the original belief state, and combination models do not update the belief state.
For example, in the second dialog turn in Fig.~\ref{sskm}, \ourmodel adds an extended triple with slot \textit{ruk} and topic ``favorite" into the belief state, while baseline DST/E2E models may update the belief state by adding triple ``restaurant-name: Pizza Hut", and combination models keep the belief state same as in the first dialog turn.
For unstructured knowledge management, we adopt standard information retrieval metrics \textbf{R@1} and \textbf{MRR@5} to evaluate the document retrieval performance.
Table \ref{table3} and \ref{table4} shows our evaluation results of belief tracking and document retrieval, respectively.

In terms of belief tracking, \ourmodel-S2S and \ourmodel-PLM outperform the non-pretrained and pretrained DST/E2E models, respectively.
This is because our extended belief tracking can detect the newly inserted turns apart from the original turns (via the slot \textit{ruk}), which improves our system's awareness on deciding when to update the triples related to original dialog process.
\ourmodel-S2S and \ourmodel-PLM also have better belief tracking performance compared to the combinations of BDA/HKS with non-pretrained and pretrained DST/E2E models, respectively.
This is because error propagation on updating belief states is eliminated in our system compared to the pipeline framework: The pipeline system either updates the belief state or retrieves the document in one turn, but \ourmodel can perform both operations in the nature of its E2E design.

In the document retrieval evaluation, we find that \ourmodel has strong performance among the unstructured knowledge management models.
In particular, \ourmodel-S2S outperforms BDA and standard IR systems, and is comparable with the strong baseline models proposed in the first track of DSTC9, which all utilize large-scale model pretraining.
While \ourmodel-PLM scores close to the strongest baseline model HKS on MRR@5 metric, and achieves the highest R@1 rate among all the baseline models.
Above experimental results indicate that our system's document retrieval scheme with topic matching has advantages over the retrieval schemes of baseline models.
Specifically, \ourmodel retrieves documents based on the highly simplified semantic information, i.e. the topic, which reduces the complexity of the retrieval process.
This makes the retrieval scheme of \ourmodel more concise and effective than the baseline models which all directly calculate the relevance of dialog context to every document content.

\subsection{Single vs. Multiple Decoders}
In this section, we compare the two extended belief state decoding schemes of our Seq2Seq system implementation \ourmodel-S2S.
We calculate the vocabularies of DSV triples, the topic and their combination (which are 709, 166 and 862), and observe that the last one approximately equals to the sum of the former two.
This confirms our assumption in Sec.~\ref{sec:ebsd} that DSV triples used for structured data query and the topic used for unstructured document retrieval have quite different vocabularies, which motivates our proposal of the multi-decoder belief state decoding scheme.

However, we find that \ourmodel-S2S (Single) outperforms \ourmodel-S2S (Multiple) in both E2E and knowledge management evaluations, as shown in Table \ref{table1}, \ref{table3} and \ref{table4}.
This shows that the decoding of DSV triples and topic can benefit from the joint optimization via shared parameters, although they are grounded on quite different vocabularies.
The superiority of joint optimization further implies that the structured and unstructured knowledge management in TOD modeling have a positive correlation, since they commonly involve task-specific domain knowledge and entities.
Therefore, the two kinds of knowledge management can learn from each other through joint training, and achieve overall better performance compared to separating them apart.
This conclusion also makes it reasonable for our system to fuse the management of structured and unstructured knowledge via their shared domains and entities.

\begin{table*}[t]
\caption{Ablation study and comparison results on knowledge management and end-to-end performances.}
\label{table5}
\centering
\resizebox{1.0\textwidth}{!}{
\smallskip\begin{tabular}{@{~}lc@{~~}ccc@{~~}c@{~~}c@{~~}c@{~~}c@{~~}c@{~}}
\hline
\multirow{2}*{Model}   & \multicolumn{2}{c}{Document Retrieval}  &  DST  &  \multicolumn{6}{c}{End-to-end}   \\
                        \cmidrule(lr){2-3}   \cmidrule(lr){4-4}   \cmidrule(lr){5-10}
                       &    MRR@5    &     R@1     & Joint Goal  &  Inform  &  Success  &   BLEU   &  METEOR  &  ROUGE-L  & Combined    \\
\hline
\ourmodel-S2S (Single) &    91.4     &    89.9     &    49.1     &   82.9   &    68.7   &   19.0   &   38.6   &    40.8   &   94.8      \\
~~~- w/o KM Fusion     & 84.0 (-7.4) & 81.7 (-8.2) & 49.1 (-0.0) &   82.9   &    68.7   &   18.3   &   37.8   &    40.2   & 94.1 (-0.7) \\
~~~- w/o Joint Optim   & 82.2 (-9.2) & 79.4 (-10.5)& 46.9 (-2.2) &   79.2   &    65.6   &   18.5   &   38.1   &    39.6   & 90.9 (-3.9) \\
~~~- repl. UT w/ GRU   & 89.7 (-1.7) & 87.9 (-2.0) & 48.0 (-1.1) &   81.9   &    68.3   &   19.3   &   39.0   &    41.2   & 94.4 (-0.4) \\
~~~- w/o F\&T (HyKnow)   & 81.7 (-9.7) & 80.2 (-9.7) & 48.0 (-1.1) &   81.9   &    68.3   &   19.0   &   38.5   &    40.9   & 94.1 (-0.7) \\
\hline
\ourmodel-PLM          &    94.7     &    93.4     &    58.5     &   93.6   &    71.9   &   17.3   &   36.8   &    40.0   &   100.1     \\
~~~- w/o KM Fusion     & 91.0 (-3.7) & 88.9 (-4.5) & 58.5 (-0.0) &   93.6   &    71.9   &   16.5   &   36.1   &    39.5   & 99.3 (-0.8) \\
~~~- w/o Joint Optim   & 90.2 (-4.5) & 86.9 (-6.5) & 57.1 (-1.4) &   92.4   &    70.3   &   16.7   &   36.0   &    39.2   & 98.1 (-2.0) \\
~~~- w/o Cons Detect   & 93.7 (-1.0) & 91.8 (-1.6) & 57.2 (-1.3) &   92.2   &    69.3   &   16.8   &   36.2   &    39.2   & 97.6 (-2.5) \\
~~~- w/~ Prev State    & 92.6 (-2.1) & 91.0 (-2.4) & 54.8 (-3.7) &   91.8   &    68.6   &   17.0   &   36.2   &    39.0   & 97.2 (-2.9) \\
\hline
\end{tabular}
}
\end{table*}

\begin{table*}[t]
\caption{More ablation study and comparison results on extended part of belief tracking and relevant entity matching.}
\label{table6}
\centering
\resizebox{0.9\textwidth}{!}{
\smallskip\begin{tabular}{lcccccccc}
\hline
\multirow{2}*{Model} & \multicolumn{3}{c}{Triple with \textit{ruk}} & \multicolumn{3}{c}{Topic} & \multicolumn{2}{c}{Entity Matching} \\
                       \cmidrule(lr){2-4}         \cmidrule(lr){5-7}       \cmidrule(lr){8-9}
                       &   P  &   R  &      F1     &  P   &  R   &      F1     &    MRR@5    &     R@1     \\
\hline
\ourmodel-S2S (Single) & 98.9 & 81.4 &     89.3    & 99.5 & 88.2 &     93.5    &     94.0    &     91.5    \\
~~~- w/o KM Fusion     & 98.8 & 81.2 & 89.1 (-0.2) & 99.5 & 87.9 & 83.3 (-0.2) & 88.6 (-5.4) & 84.9 (-6.6) \\
\ourmodel-PLM          & 99.4 & 88.0 &     93.4    & 99.6 & 94.1 &     96.8    &     96.4    &     94.4    \\
~~~- w/o KM Fusion     & 99.4 & 87.9 & 93.3 (-0.1) & 99.6 & 94.1 & 96.8 (-0.0) & 93.3 (-3.1) & 90.5 (-3.9) \\
\hline
\end{tabular}
}
\end{table*}

\begin{table*}[t]
\caption{More ablation study and comparison results on computational efficiency.}
\label{table7}
\centering
\resizebox{0.75\textwidth}{!}{
\smallskip\begin{tabular}{lcccc}
\hline
         Model         & Model Size & Training Time & Inference Time & Memory Usage \\
\hline
\ourmodel-S2S (Single) &    43.7M   &    368mins    &     0.026s     &       9GB    \\
~~~- repl. UT w/ GRU   &    4.08M   &    767mins    &     0.107s     &      11GB    \\
\ourmodel-PLM          &    124M    &    714mins    &     1.296s     &      37GB    \\
\hline
\end{tabular}
}
\end{table*}

\subsection{Ablation Study and More Comparisons}
In this section, we conduct ablation study and performance comparisons on our two system implementations \ourmodel-S2S (Single) and \ourmodel-PLM.

We consider to ablate \textbf{1) the fusion of structured and unstructured knowledge management (KM)}, or further \textbf{2) the joint optimization of structured and unstructured knowledge-grounded TOD modeling}, to investigate their respective roles in our system, denoted as \textbf{w/o KM Fusion} and \textbf{w/o Joint Optim}.
Both ablation 1) and 2) separate apart the structured and unstructured knowledge, i.e. we split the semi-structured knowledge base back into the original database and document base in Mod-MultiWOZ 2.1.
Moreover, we \textbf{3) replace the Universal Transformers (UT) with GRU networks in \ourmodel-S2S (Single)}, to compare the Seq2Seq modeling ability and efficiency of UT and GRU in semi-structured knowledge-grounded TOD, denoted as \textbf{repl. UT w/ GRU}.
By applying \textbf{both 1) and 3)}, denoted as \textbf{w/o F\&T}, we degrade \ourmodel-S2S to our prior work HyKnow \cite{gao2021hyknow}, to further show the effectiveness of our system improvements.
For \ourmodel-PLM, we also ablate \textbf{4) the consistency detection} to verifies its benefits to our system, denoted as \textbf{w/o Cons Detect}.
Besides, we train \ourmodel-PLM to \textbf{5) decode extended belief state $\widetilde{B}_{t}$ based on previous state $\widetilde{B}_{t-1}$ and short context $C_{t}=[R_{t-1},U_{t}]$}, to verify our statement that directly decoding $\widetilde{B}_{t}$ based on full $C_{t}$ can avoid error accumulation, denoted as \textbf{w/ Prev State}.

We compare the performance of \ourmodel-S2S (Single), \ourmodel-PLM and their ablation models from multiple aspects.
Specifically, we follow Sec.~\ref{sec:e2e} and \ref{sec:km} to compare the E2E and knowledge management performances of \ourmodel-S2S (Single), \ourmodel-PLM and their ablation models, whose results are shown in Table \ref{table5}.
In the ablation of KM Fusion, we also evaluate the extended part of belief tracking and relevant entity matching performances to further investigate the management of unstructured knowledge, whose results are shown in Table \ref{table6}.
Specifically, we evaluate the \textbf{P}recision, \textbf{R}ecall and \textbf{F1}-measure of predicting the triple with \textit{ruk} and the topic in our extended belief state.
And we follow the document retrieval evaluation to compare the \textbf{MRR@5} and \textbf{R@1} rates in matching the relevant entity whose unstructured documents are involved.
Besides, we compare the computational efficiency of \ourmodel-S2S (Single), \ourmodel-S2S (Single) repl. UT w/ GRU and \ourmodel-PLM, including model size, total training time, average inference time per dialog turn and memory usage, whose results are shown in Table \ref{table7}.

\subsubsection{w/o KM Fusion}
We use the original and extended parts of belief state to separately perform structured database query and unstructured document retrieval, where the value of slot \textit{ruk} and the topic are used to retrieve the documents of all entities in the active domain.
Without knowledge management fusion, we observe that \ourmodel suffers from evident performance declines in terms of document retrieval, which leads to lower language quality of generated responses.
Meanwhile, \ourmodel's MRR@5 and R@1 rates in matching the relevant entity also dramatically decrease, although the F1-measure of the belief state's extended part (i.e. the triple with \textit{ruk} and the topic) varies much less.
The above results indicate that knowledge management fusion brings great benefits on finding relevant entities and documents.
Specifically, through the knowledge management fusion, the original triples in the belief state can provide more constraints via structured query to help narrow down the candidates of relevant entities and documents.
This simplifies the matching of entity name/ID and topic in dialog turns requiring unstructured knowledge, and avoids some document retrieval errors caused by the mis-prediction of belief state's extended part.

\subsubsection{w/o Joint Optim}
We train two \ourmodel models to address our TOD modeling task: one tracks the original belief state, performs database query and generates responses in original dialog turns, while the other parses the extended part of belief state, performs document retrieval and generates responses in newly inserted dialog turns.
In testing, we use the absence or presence of slot \textit{ruk} to judge whether a dialog turn belongs to original or newly inserted turns.
We observe that removing joint optimization brings \ourmodel evident performance declines in the E2E evaluation. This suggests that joint optimization plays a significant role in improving \ourmodel's E2E performance, where TOD modeling grounded on structured and unstructured knowledge can benefit each other by learning shared parameters.
The ablation of joint optimization also causes further declines in \ourmodel's knowledge management performance, compared to that without KM Fusion. This again indicates that structured and unstructured knowledge management are positively correlative and can get benefit from joint training.

\subsubsection{repl. UT w/ GRU}
We use bidirectional GRU as encoders and standard GRU as decoders to replace Universal Transformers in \ourmodel-S2S (Single).
After the replacement, the knowledge management performance of \ourmodel-S2S (Single) decreases, which also causes the lower task completion rate (i.e. Inform and Success rate) in E2E evaluation.
The above results prove that Universal Transformers has stronger Seq2Seq modeling ability in semi-structured knowledge-grounded TOD, compared to GRU networks.
In terms of computational efficiency, we find that under similar memory usage, \ourmodel-S2S (Single) developed on Universal Transformers requires less training time than that developed on GRU, although the former has much larger model size than the latter.
Besides, \ourmodel-S2S (Single) developed on Universal Transformers has faster inference speed.
These results show that Universal Transformers also has higher efficiency than GRU in terms of semi-structured knowledge-grounded TOD modeling.

\subsubsection{w/o F\&T}
By removing both knowledge management fusion and Universal Transformers (replaced by GRU), SeKnow-S2S is degraded to our prior work HyKnow.
We observe that SeKnow-S2S outperforms HyKnow in both knowledge management and end-to-end performances, especially in terms of document retrieval, which further shows that our system improvements are effective.

\subsubsection{w/o Cons Detect}
We remove the auxiliary consistency detection task in \ourmodel-PLM, which results in performance decline in both knowledge management and end-to-end evaluation.
This indicates that consistency detection still has effectiveness in improving TOD modeling performance when both structured and unstructured knowledge are involved.

\subsubsection{w/ Prev State}
When decoding belief state with previous state and short context, the joint goal accuracy of SeKnow-PLM decreases dramatically, which also brings down the performances of SeKnow-PLM in other aspects.
This proves our statement that directly decoding belief state based on full context can avoid error accumulation.

\subsubsection{\ourmodel-S2S vs. \ourmodel-PLM}
With the usage of pretrained language model, SeKnow-PLM achieves significantly better performances than \ourmodel-S2S (Single) in terms of extended belief tracking, document retrieval and task completion.
In trade-off, \ourmodel-PLM has a larger model size and requires more memory to get comparable training time with \ourmodel-S2S (Single).
The inference speed of \ourmodel-PLM is also much slower than \ourmodel-S2S (Single).
This shows that large-scale model pretraining has great power in improving TOD modeling grounded on semi-structured knowledge, while at the cost of extra pretraining corpus and lower computational efficiency.
Noting that although \ourmodel-PLM does not score higher than \ourmodel-S2S (Single) on BLEU, METEOR and ROUGE-L, the three metrics can not absolutely represent the language quality of system response.
Through the human evaluation in Sec.~\ref{sec:he}, we prove that \ourmodel-PLM actually has better response quality than our non-pretrained model \ourmodel-S2S (Single).

\begin{table*}[t]
\caption{Belief tracking and end-to-end evaluation results on the original MultiWOZ 2.1 and Mod-MultiWOZ 2.1 test set. The evaluation is conducted only in the original dialog turns.}
\label{table8}
\centering
\resizebox{1.0\textwidth}{!}{
\smallskip\begin{tabular}{clccccccc}
\hline
        Test Set          & Model                  &  Joint Goal  & Inform & Success &  BLEU  & METEOR & ROUGE-L &  Combined  \\
\hline
\multirow{4}{*}{MultiWOZ 2.1} & LABES-S2S + HKS        &     49.3     &  82.3  &  69.7   &  17.8  &  37.1  &  40.2   &    93.8    \\
                          & AuGPT + HKS            &     58.5     &  93.4  &  73.5   &  17.2  &  36.5  &  40.0   &   100.7    \\
                          & \ourmodel-S2S (Single) &     49.9     &  83.5  &  69.3   &  18.9  &  38.2  &  41.0   &    95.3    \\
                          & \ourmodel-PLM          &     58.9     &  95.6  &  72.4   &  17.9  &  36.8  &  40.2   &   101.9    \\
\hline
\multirow{4}{*}{Mod-MultiWOZ 2.1} & LABES-S2S + HKS        &  47.6 (-1.7) &  78.2  &  66.7   &  17.6  &  36.8  &  39.6   &  90.1 (-3.7) \\
                          & AuGPT + HKS            &  56.6 (-1.9) &  91.6  &  70.7   &  16.9  &  36.3  &  39.3   &  98.1 (-2.6) \\
                          & \ourmodel-S2S (Single) &  49.1 (-0.8) &  82.9  &  68.7   &  18.2  &  37.3  &  40.5   &  94.0 (-1.3) \\
                          & \ourmodel-PLM          &  58.5 (-0.4) &  93.6  &  71.9   &  17.5  &  36.6  &  40.0   & 100.3 (-1.6) \\
\hline
\end{tabular}
}
\end{table*}

\subsection{Between Structured and Unstructured Knowledge}
In this section, we investigate how the newly inserted dialog turns (involving unstructured knowledge) affect systems' TOD performances (i.e. tracking user goals associated with structured knowledge and generating responses) in the original dialog turns.
Specifically, we evaluate systems' belief tracking and E2E performances on both the original MultiWOZ 2.1 and Mod-MultiWOZ 2.1 test sets.
This evaluation is conducted only in the original dialog turns, which is different from the E2E evaluation conducted in all turns (Table \ref{table1}).
We evaluate our non-pretrained and pretrained system implementations \ourmodel-S2S (Single) and \ourmodel-PLM, compared with their corresponding strong baseline models LABES-S2S + HKS and AuGPT + HKS.
The results of this experiment are shown in Table \ref{table8}, where both comparisons come to the same conclusions described as below.

We first find that all the models' performances are degraded when transferred from MultiWOZ to Mod-MultiWOZ 2.1 test set. This indicates that the newly inserted turns involving new domain knowledge may interrupt the original dialogs, which complicates the dialog process and causes difficulties in the original turns' dialog modeling.

However, we observe that \ourmodel suffers from less reduction compared to the baseline combination models.
This shows that our system has a stronger resistance to the interruptions of newly inserted dialog turns, which benefits from our end-to-end modeling.
Specifically, \ourmodel jointly optimizes dialog modeling of the original and newly inserted dialog turns in a uniform end-to-end framework. This unified modeling approach improves our system's flexibility in switching between the two kinds of turns, and thus makes it more competent in handling the complicated dialog process.

\begin{table}[t]
\caption{Human evaluation results on Mod-MultiWOZ 2.1, results in original and newly inserted turns are shown separately.}
\label{table9}
\centering
\resizebox{1.0\columnwidth}{!}{
\smallskip\begin{tabular}{@{~}l@{~~}c@{~}c@{~}c@{~~}c@{~}c@{~}c@{~}}
\hline
\multirow{2}{*}{Model} & \multicolumn{3}{c}{Original}                  & \multicolumn{3}{c}{Newly Inserted}            \\
                         \cmidrule(r){2-4}                              \cmidrule(r){5-7}              
                       &      Cohe.    &      Info.    &     Corr.     &      Cohe.    &      Info.    &      Corr.    \\
\hline
\ourmodel-S2S (Single) &      2.60     &      2.56     &      2.48     &      2.58     &      2.50     &      2.54     \\
\hline
AuGPT                  &      2.66     &      2.68     &      2.60     &      2.62     &      2.54     &      2.58     \\
AuGPT + HKS            & \textbf{2.68} &      2.66     &      2.64     &      2.64     &      2.66     &      2.62     \\
\ourmodel-PLM          &      2.64     & \textbf{2.70} & \textbf{2.68} & \textbf{2.66} & \textbf{2.70} & \textbf{2.72} \\
\hline
\end{tabular}
}
\end{table}

\subsection{Human Evaluation}\label{sec:he}
There is still a gap between the evaluation results of automatic metrics and the real E2E performance of TOD systems.
Therefore, we conduct human evaluation to more adequately test our system's E2E performance.
In particular, we test the performance of \ourmodel-PLM which achieves the best combined score in the automatic E2E evaluation, compared with the strongest E2E baseline AuGPT and its combination with HKS.
We also compare the above models with our non-pretrained system implementation \ourmodel-S2S (Single), to further investigate the effectiveness of model pretraining in E2E TOD modeling.

We conduct human evaluation separately on the two types (original and newly inserted) of dialog turns.
Specifically, we first sample fifty dialog turns of each type, combined with their dialog context in previous turns.
Based on our sampled dialog segments, we then design an online questionnaire and hire English native speakers to finish it.
The questionnaire contains three types of tasks, which instruct the participants to evaluate the generated response on three aspects:
\begin{itemize}
    \item \textbf{Coherence (Cohe.)}: Participants are asked to judge whether the generated response is coherent with the given dialog context, by scoring on a Liker scale of 1 (not coherent), 2 (roughly coherent), and 3 (coherent).
    \item \textbf{Informativeness (Info.)}: Participants are asked to judge whether the generated response contains information requested by the user in the given dialog context, by scoring on a Liker scale of 1 (lack of information), 2 (barely enough information), and 3 (sufficient information). Ground truth response is given as a reference contrast, which is supposed to contain sufficient information.
    \item \textbf{Correctness (Corr.)}: Participants are asked to judge whether the information in generated response is consistent with the given ground truth knowledge (relevant database entries or documents), by scoring on a Liker scale of 1 (not consistent), 2 (partly consistent), and 3 (totally consistent).
\end{itemize}
Table \ref{table9} shows our human evaluation results.

We find that \ourmodel-PLM, AuGPT and AuGPT + HKS significantly outperform \ourmodel-S2S (Single) in both original and newly inserted dialog turns, especially in terms of informativeness and correctness.
This indicates that large-scale model pretraining has great benefits in promoting the response quality in E2E TOD modeling, although pretrained LM does not bring higher scores of language quality (i.e. BLEU, METEOR and ROUGE-L) in automatic E2E evaluation (Table \ref{table1}).

Compared to AuGPT and AuGPT + HKS, \ourmodel-PLM scores higher on informativeness and correctness in the original dialog turns.
The p-values of paired sample t-test between \ourmodel-PLM and AuGPT + HKS on informativeness and correctness are 0.042 and 0.038, respectively.
This is consistent with the automatic evaluations, showing that \ourmodel-PLM has better belief tracking and task completion performances.
\ourmodel-PLM also outperforms AuGPT and AuGPT + HKS in the newly inserted dialog turns.
Specifically, \ourmodel-PLM generates responses with significantly better informativeness and correctness than AuGPT, with p-values 0.007 and 0.009 in paired sample t-test.
This again shows that the management of unstructured knowledge is beneficial for generating appropriate responses.
Compared to AuGPT + HKS, the responses generated by \ourmodel-PLM also achieve much better correctness, with p-value 0.013 in paired sample t-test. This benefits from our model’s higher document retrieval accuracy (i.e. higher R@1 rate in Table \ref{table4}).

\begin{table*}[t]
\caption{Comparisons of belief tracking, document retrieval and response generation results when dialog steps into a turn involving unstructured knowledge. ``Human" denotes ground truth results.}
\label{table10}
\centering
\small
\resizebox{1.0\textwidth}{!}{
\smallskip\begin{tabular}{c|l@{~}l}
\hline
\multicolumn{3}{l}{$U_{1}$: We are looking for \textbf{free wifi} and a nice \textbf{4 star} \textbf{guesthouse} that has \textbf{free parking}. We do not care about area or price range.} \\
\multicolumn{3}{l}{$B_{1}$: hotel-internet: yes, hotel-parking: yes, hotel-stars: 4, hotel-type: guesthouse} \\
\multicolumn{3}{l}{$R_{1}$: I would recommend the \textbf{Acorn Guest House}. It is a moderately priced guesthouse in the north. Would you like me to book a room?} \\
\hline
\multicolumn{3}{l}{$U_{2}$: Do they provide an \textbf{\textcolor{blue}{Italian} breakfast}?} \\
\hline
\multirow{3}*{Human} & $B_{2}$: & hotel-internet: yes, hotel-parking: yes, hotel-stars: 4, hotel-type: guesthouse \\
                     & $D_{2}$: & Continental, vegetarian, and a full English and Irish \textbf{breakfast} are available at \textbf{Acorn Guest House}. \\
                     & $R_{2}$: & \textbf{No}, they do not but they have Irish breakfast. Would you like me to go ahead and book it? \\
\hline
\multirow{2}*{AuGPT} & $B_{2}$: & \textbf{restaurant-food: \textcolor{blue}{Italian}}, hotel-internet: yes, hotel-parking: yes, hotel-stars: 4, hotel-type:                          guesthouse \\
                     & $R_{2}$: & \textbf{Da Vinci Pizzeria} is an \textbf{\textcolor{blue}{Italian} restaurant} in the north. Would you like to book a table? \\
\hline
\multirow{3}*{AuGPT + HKS} & $B_{2}$: & hotel-internet: yes, hotel-parking: yes, hotel-stars: 4, hotel-type: guesthouse \\
                           & $D_{2}$: & German, \textbf{\textcolor{blue}{Italian}}, Spanish and English are \textbf{spoken} at Carolina Bed and \textbf{Breakfast}. \\
                           & $R_{2}$: & \textbf{Yes}, they offer both Italian and English breakfast. Is there anything else I can help you with? \\
\hline
\multirow{4}*{\ourmodel-S2S (Single)} & \multirow{2}*{$\widetilde{B}_{2}$:} & hotel-internet: yes, hotel-parking: yes, hotel-stars: 4, hotel-type: guesthouse, \\
                    &          & \textbf{hotel-ruk: guesthouse}, \textbf{topic: breakfast option} \\
                    & $D_{2}$: & Continental, vegetarian, and a full English and Irish \textbf{breakfast} are available at \textbf{Acorn Guest House}. \\
                    & $R_{2}$: & \textbf{No}, they have a full English and Irish breakfast. Would you like to make a reservation? \\
\hline
\multirow{4}*{\ourmodel-PLM}   & \multirow{2}*{$\widetilde{B}_{2}$:} & hotel-internet: yes, hotel-parking: yes, hotel-stars: 4, hotel-type: guesthouse, \\
                    &          & \textbf{hotel-ruk: Acorn Guest House}, \textbf{topic: breakfast option} \\
                    & $D_{2}$: & Continental, vegetarian, and a full English and Irish \textbf{breakfast} are available at \textbf{Acorn Guest House}. \\
                    & $R_{2}$: & \textbf{No}, they provide Continental, vegetarian, and a full English and Irish breakfast. Anything else I can do for you? \\
\hline
\multicolumn{3}{l}{$U_{2}$: Do they provide an \textbf{\textcolor{red}{Irish} breakfast}?} \\
\hline
\multirow{3}*{Human} & $B_{2}$: & hotel-internet: yes, hotel-parking: yes, hotel-stars: 4, hotel-type: guesthouse \\
                     & $D_{2}$: & Continental, vegetarian, and a full English and \textbf{\textcolor{red}{Irish} breakfast} are available at \textbf{Acorn Guest House}. \\
                     & $R_{2}$: & \textbf{Yes}, they provide Irish breakfast. Would you like me to go ahead and book it? \\
\hline
\multirow{2}*{AuGPT} & $B_{2}$: & \textbf{restaurant-food: \textcolor{red}{Irish}}, hotel-internet: yes, hotel-parking: yes, hotel-stars: 4, hotel-type:                          guesthouse \\
                     & $R_{2}$: & Sorry, there are no \textbf{\textcolor{red}{Irish} restaurants} like that. Would you like to change the type of food? \\
\hline
\multirow{3}*{AuGPT + HKS} & $B_{2}$: & hotel-internet: yes, hotel-parking: yes, hotel-stars: 4, hotel-type: guesthouse \\
                           & $D_{2}$: & Continental, vegetarian, and a full English and \textbf{\textcolor{red}{Irish} breakfast} are available at \textbf{Acorn Guest House}. \\
                           & $R_{2}$: & \textbf{Yes}, they offer continental, vegetarian, and a full English and Irish breakfast. Is there anything else I can help? \\
\hline
\multirow{4}*{\ourmodel-S2S (Single)} & \multirow{2}*{$\widetilde{B}_{2}$:} & hotel-internet: yes, hotel-parking: yes, hotel-stars: 4, hotel-type: guesthouse, \\
                    &          & \textbf{hotel-ruk: guesthouse}, \textbf{topic: breakfast option} \\
                    & $D_{2}$: & Continental, vegetarian, and a full English and \textbf{\textcolor{red}{Irish} breakfast} are available at \textbf{Acorn Guest House}. \\
                    & $R_{2}$: & \textbf{Yes}, they have a full English and Irish breakfast. Would you like to book a room? \\
\hline
\multirow{4}*{\ourmodel-PLM}   & \multirow{2}*{$\widetilde{B}_{2}$:} & hotel-internet: yes, hotel-parking: yes, hotel-stars: 4, hotel-type: guesthouse, \\
                    &          & \textbf{hotel-ruk: Acorn Guest House}, \textbf{topic: breakfast option} \\
                    & $D_{2}$: & Continental, vegetarian, and a full English and Irish \textbf{\textcolor{red}{Irish} breakfast} are available at \textbf{Acorn Guest House}. \\
                    & $R_{2}$: & \textbf{Yes}, they provide continental, vegetarian, and a full English and Irish breakfast. Anything else I can do? \\
\hline
\end{tabular}
}
\end{table*}


\subsection{Case Study}
In this section, we investigate a case of TOD when dialog steps into a newly inserted turn of Mod-MultiWOZ 2.1, involving unstructured knowledge.
The example dialog segment ($U_1$, $B_1$, $R_1$, $U_2$) and corresponding output results of each model ($B_2/\widetilde{B}_{2}$, $D_2$, $R_2$) are presented in Table \ref{table10}.
To give more adequate analysis, we also test each model's performance with a modified input case, by replacing the key word ``Italian" in $U_2$ with another food type ``Irish".

Without utilization of unstructured knowledge, AuGPT misunderstands the user query in both original and modified cases, and recognizes the word ``Italian"/``Irish" in $U_2$ as a constraint to update the belief state. As a result, the system makes inappropriate recommendations.

By combining with HKS, AuGPT predicts correct belief state, but fails in finding the relevant document in original case, thus providing a wrong answer.
This is because the wrong document’s content has many common words with the original dialog context, e.g. ``Italian" and ``Breakfast", which mislead the retrieval process.
After replacing ``Italian" with ``Irish", the misleading is eliminated, and AuGPT + HKS finds the correct document and gives an appropriate response.

In both original and modified cases, \ourmodel-S2S (Single) successfully finds the relevant document, although failing to predict the relevant entity in $\widetilde{B}_{2}$. This is because \ourmodel-S2S (Single) uses the original triples in $\widetilde{B}_{2}$ to help locate the relevant entity, which avoids the document retrieval error caused by the vague prediction of \textit{ruk}'s value.

With the power of large-scale model pretraining, \ourmodel-PLM does not get confused by the interference words in dialog context. Therefore, it successfully identifies the relevant entity (``Acorn Guest House") and topic (``breakfast option"), and generates proper responses with accurate information in both original and modified input cases.

\begin{table}[t]
\caption{Evaluation results on original MultiWOZ 2.1 dataset.}
\label{table11}
\centering
\resizebox{1.0\columnwidth}{!}{
\smallskip\begin{tabular}{@{~}l@{~}c@{~~}c@{~~}c@{~~}c@{~~}c@{~}}
\hline
Model                  & Joint Goal  &    Inform   &   Success   &    BLEU     &   Combined  \\
\hline
UniConv                &     50.1    &     72.6    &     62.9    &\textbf{19.8}&     87.6    \\
LABES-S2S              &     51.5    &     78.1    &     67.1    &     18.1    &     90.7    \\
\ourmodel-S2S (Single) &\textbf{52.0}&\textbf{79.2}&\textbf{67.9}&     18.9    &\textbf{92.5}\\
\hline
MinTL                  &     53.6    &     84.9    &\textbf{74.9}&\textbf{17.9}&      97.8     \\
AuGPT                  &     57.0    &     91.4    &     72.9    &     17.2    &      99.4     \\
\ourmodel-PLM          &\textbf{57.8}&\textbf{91.8}&     73.4    &     17.8    &\textbf{100.4} \\
\hline
\end{tabular}
}
\end{table}

\begin{table}[t]
\caption{Evaluation results on DSTC9 Track1 dataset.}
\label{table12}
\centering
\resizebox{1.0\columnwidth}{!}{
\smallskip\begin{tabular}{@{~}l@{~}c@{~~}c@{~~}c@{~~}c@{~~}c@{~}}
\hline
Model                  &    MRR@5    &     R@1     &    BLEU-1   &    METEOR   &    ROUGE-L  \\
\hline
BDA                    &     72.6    &     62.0    &     30.3    &     29.8    &     30.4    \\
E2E-DGC                &     92.3    &     89.6    &     35.2    &     35.3    &     35.0    \\
DKR                    &     90.3    &     87.2    &     35.2    &     37.0    &     35.6    \\
HKS                    &\textbf{94.5}&\textbf{93.2}&\textbf{37.9}&\textbf{38.6}&     37.1    \\
\hline
\ourmodel-S2S (Single) &     87.8    &     84.4    &     33.6    &     34.2    &     34.0    \\
\ourmodel-PLM          &     93.5    &     91.0    &     36.8    &     38.0    &\textbf{37.2} \\
\hline
\end{tabular}
}
\end{table}

\subsection{Original MultiWOZ 2.1 and DSTC9 Track1 Evaluation}
We also investigate whether \ourmodel still has strong performance in TOD modeling when only structured/unstructured knowledge is available.
In particular, we evaluate the performance of \ourmodel and baseline models on original MultiWOZ 2.1 and DSTC9 Track1 task, whose experimental results are shown in Table \ref{table11} and \ref{table12}, respectively.

On original MultiWOZ 2.1 dataset, we observe that \ourmodel-S2S (Single) and \ourmodel-PLM have comparable DST/E2E evaluation results with strong non-pretrained and pretrained baseline models, respectively.
This shows that \ourmodel still maintains strong performance in traditional TOD modeling grounded only on structured knowledge.

In DSTC9 Track1 task, the system needs to modeling TOD with access to the document base in Mod-MultiWOZ 2.1 dataset, where structured knowledge and belief state labels are not available.
Instead of generating the whole extended belief state, we train \ourmodel to directly decode relevant entity (i.e. value of slot \textit{ruk}) and topic, which are then used for the subsequent document retrieval and response generation.
We find that \ourmodel-PLM achieves comparable evaluation results with the strongest baseline HKS, and even non-pretrained \ourmodel-S2S (Single) scores close to the pretrained baseline models.
This indicates that \ourmodel also has a strong performance in TOD modeling grounded only on unstructured knowledge.

\section{Conclusion}
In this paper, we define a task of modeling TOD with management of semi-structured knowledge.
To address this task, we propose a TOD system \ourmodel and introduce its two implementations, one (\ourmodel-PLM) with and one (\ourmodel-S2S) without model pretraining.
Both implementations use an extended belief tracking to manage semi-structured knowledge, and jointly optimize TOD modeling grounded on structured and unstructured knowledge in the E2E manner.
In the experiments, \ourmodel shows strong performance in TOD modeling with semi-structured knowledge management, compared to existing TOD systems and their pipeline extensions.
For future work, we consider evaluating our system on more various TOD scenarios where dialogs are grounded on semi-structured knowledge.



\appendices
\section{Document Preprocessing}
\label{document}
We preprocess the documents of each entity in the knowledge base of Mod-MultiWOZ 2.1 \cite{kim2020beyond} dataset to extract the topic of each document, used as its retrieval index in the semi-structured knowledge management.
Based on the TF-IDF \cite{manning2008introduction} algorithm, we perform the topic word extraction domain-by-domain in a two-step procedure.
First, we choose the top-three keywords with the highest TF-IDF scores in each document as its topic candidates.
Then we filter the candidates to further select our desired topic words.

Noticing that different entities in the same domain usually have documents covering similar topics, we assume that a desired topic word should typically appear in multiple entities' documents, and therefore have a high frequency of occurrence among the topic candidates.
So we calculate a cumulative average TF-IDF (CA-TF-IDF) score for each topic word in the candidates, which synthetically measures the word's document-level TF-IDF and entity-level occurrence frequency.
Specifically, CA-TF-IDF sums the TF-IDF score of a topic word's each occurrence in the candidates, and divides it by the entity number in the domain.
We filter out the topic candidates with low CA-TF-IDF scores and retain the rest to form the final retrieval indexes.
The filtering thresholds are 2.3, 2.7, 6.9 and 7.3 for the domain of restaurant, hotel, taxi and train, respectively.
While other domains are not involved in the unstructured documents.
After the preprocessing, each document has one to three topic words extracted.

\section{Training and Implementation Details}
\label{implement}
In \ourmodel-S2S, we use two-layer Universal Transformers \cite{dehghani2018universal} to implement our encoders/decoders, with 8 parallel attention heads in each self-attention network.
We set the convolution kernel size and inner-layer size of feed-forward network as 3 and 2048, and the batch size, embedding/hidden size and vocabulary size as 64, 512 and 3000, respectively.
We also set dropout rate as 0.1 and use greedy decoding to generate the belief state and system response.
Moreover, we use Adam optimizer \cite{kingma2014adam} with an initial learning rate of $3e^{-4}$, and decay the learning rate by dividing it by current epoch number.
We set the total number of training epoch as 15, with average training time 25 minutes per epoch using 1 GPU.
In \ourmodel-PLM, we follow AuGPT \cite{kulhanek2021augpt} to pretrain a GPT-2 \cite{radford2019language} as our model backbone, using its suggested implementation settings\footnote{\url{https://huggingface.co/jkulhanek/augpt-bigdata/tree/main}}.
We further finetune the GPT-2 on the training set of Mod-MultiWOZ 2.1, where we set the batch size as 8.
We set the total number of finetuning epoch as 7, with average training time 100 minutes per epoch using 4 GPUs.
Model training is performed on NVIDIA TITAN-Xp GPU.
Besides, in semi-structured knowledge operation, we conduct the matching of entity name/ID and topic by using the fuzzy string matching toolkit\footnote{\url{https://github.com/seatgeek/fuzzywuzzy}}. Specifically, we first calculate the simple fuzzy match ratio (i.e. fuzz.ratio) between slot \textit{ruk}’s value and each entity’s name/ID, and select entity with the highest ratio as the best-matched one. For each document of the best-matched entity, we then calculate the token sorted fuzzy match ratio (i.e. fuzz.token\_sort\_ratio) between its topic and generated topic ($T_{t}$), and select document with the highest ratio as the relevant one ($D_{t}$).

\section{Statistics of Mod-MultiWOZ 2.1}
\label{statistics}
There are totally 8449/1001/1004 dialogs\footnote{These are slightly more compared to the original MultiWOZ 2.1, because some of the original dialogs are modified twice with different turns inserted.} in the training, development and testing set of Mod-MultiWOZ 2.1, where 6501/836/847 dialogs have new turns inserted, respectively.
After the modification, each dialog has 8.93 turns on average, which is longer than the original 6.85.
The ontology of Mod-MultiWOZ 2.1 is same as the original, with 32 slot types (excluding \textit{ruk}) and 2426 corresponding slot values.
There are totally 291 entities in the semi-structured knowledge base: 110, 79, 66 and 33 entities in the domain of restaurant, attraction, hospital and hotel, respectively, and 1 entity in each of the other three domains\footnote{In the domain of train or taxi, we assume that different train schedules or taxi cars belong to a common entity.} (i.e. police, train and taxi).
Entities in the domain of restaurant, hotel, train and taxi are associated with unstructured documents, whose total number is 2882.

\section*{Acknowledgment}
This work was supported by the National Science Foundation for Distinguished Young Scholars (with No. 62125604) and the NSFC projects (Key project with No. 61936010 and regular project with No. 61876096).
This work was also supported by the Guoqiang Institute of Tsinghua University, with Grant No. 2019GQG1 and 2020GQG0005.
This work was also supported by the Allen Institute for AI.
We would also like to thank the anonymous reviewers for their invaluable suggestions and feedback.

\ifCLASSOPTIONcaptionsoff
  \newpage
\fi



\bibliographystyle{IEEEtran}
\bibliography{main}

\end{document}